\documentclass[journal]{IEEEtran}

\usepackage{cite}  
\usepackage{url} 
\usepackage{amsmath,amssymb,amsfonts} 
\usepackage{float}
\newfloat{Equation}{H}{equ}[section]
\floatname{Equation}{Equation}
\usepackage{graphicx}
\usepackage{booktabs}
\usepackage{enumitem}
\usepackage{textcomp}
\usepackage[hidelinks]{hyperref}
\usepackage{placeins}
\usepackage{stfloats}   
\usepackage{adjustbox}

\title{Real-Time Obstacle Detection and Distance Estimation on Smartphones for Visually Impaired and Distracted Pedestrians}

\author{Abtin~Zandi,~Sarvin~Nami,~Roza~Ganjipour,~Bardia~Sabagh~Kermani,~Amirreza~Tarabkhah,~and~Hamed~Farbeh%

\thanks{Abtin Zandi, Sarvin Nami, Roza Ganjipour, Bardia Sabagh Kermani, and Amirreza Tarabkhah are with the Department of Computer Engineering, Amirkabir University of Technology, Tehran, Iran. (e-mail: abtinzandi@gmail.com,srvn0nm@gmail.com, rozaganjipour@aut.ac.ir, bardia112@aut.ac.ir, tarabkhah2@gmail.com).}%

\thanks{Hamed Farbeh is with the Al-Khwarizmi University, Urgench 220100, Uzbekistan. (e-mail: h.farbeh@akhu.uz).}%
}

\begin{document}

\maketitle

\begin{abstract}
According to the World Health Organization, at least 2.2 billion people worldwide live with visual impairment, including tens of millions with severe vision loss. Traditional aids such as white canes or guide dogs remain costly, limited in capability, and inaccessible to many. At the same time, smartphone distraction in public spaces has been linked to elevated risks of collisions with vehicles, trips, and falls. These dual challenges motivate lightweight, smartphone-based solutions that combine accessibility with low latency. We present a Kotlin-based Android application that uses a YOLOv8 detector, retrained to recognize eighteen common urban obstacle classes, to deliver real-time obstacle alerts for both user groups. The detector was optimized with pruning, quantization, and early-exit branches, and deployed through the Android Neural Networks API (NNAPI), which can dispatch inference to GPU, NPU, or DSP accelerators when available. On a mid-range smartphone, these optimizations reduced average inference latency from about 3.2 s for the off-the-shelf YOLOv8n to 0.39-0.48 s; the recommended configuration (INT8 quantization with early exit) runs in 0.42-0.48 s while retaining a recall of 0.81. The application combines the optimized detector with ARCore-based or monocular focal-length distance estimation, so it works with a single smartphone camera and requires no LiDAR, depth camera, or other specialized sensor. Multimodal feedback, including Text-to-Speech guidance for visually impaired users and vibration alerts for distracted pedestrians, supports use across different contexts.
\end{abstract}


\section{Introduction}
\IEEEPARstart{T}{he} rapid growth of smartphone usage and their wide availability in daily life have transformed communication and routine activities. However, smartphone-related distraction while walking in public spaces has introduced new safety risks. Mobile-phone distraction has been shown to reduce situational awareness and balance, increasing the likelihood of collisions with vehicles, trips, and accidental falls \cite{Stavrinos2018, Pelicioni2023}. At the same time, the global burden of visual impairment remains substantial: the World Health Organization estimates that at least 2.2 billion people experience some form of vision loss, with tens of millions facing severe conditions \cite{WHO2023, VisionLoss2021}. Conventional aids such as white canes or guide dogs, although effective, offer limited situational awareness and remain inaccessible or unaffordable for many users \cite{WHO2019}. These challenges highlight the urgent need for smartphone-based assistive systems that are widely available, operate with low latency, and improve mobility safety for both visually impaired individuals and distracted pedestrians.

Advances in artificial intelligence and machine vision have enabled new opportunities to address these challenges. State-of-the-art object-detection architectures such as YOLO have demonstrated strong performance in identifying surrounding objects in real-time. When adapted for mobile platforms, such models can provide timely information about the presence and location of obstacles, improving both safety and user experience. However, standard YOLO variants are often too computationally intensive for mid-range smartphones, leading to multi-second delays per frame that make real-time detection impractical. Many existing assistive-vision applications suffer from delayed response times, limited recognition vocabularies, or dependence on expensive sensors \cite{Tapu2013, Messaoudi2022}. 

This paper proposes a practical, real-time Android-based solution that extends the YOLOv8 architecture through lightweight optimization techniques. The proposed application targets two primary user groups: (1) individuals with visual impairments, who receive Text-to-Speech voice feedback, and (2) distracted pedestrians, who receive vibration-based alerts. The detection model was retrained to recognize eighteen obstacle categories and optimized through pruning, early-exit mechanisms, and quantization to reduce latency, and we evaluate the accuracy cost of each optimization.

Through these optimizations, the inference time of the baseline YOLOv8n (FP32) model on a Samsung M32 was reduced from approximately 3.2 s per image to approximately 0.99 s after customization, 0.55 s with INT8 quantization, and as low as 0.39–0.40 s when pruning is added, an approximately 8× speedup. Because pruning substantially reduces recall, we recommend the INT8 + early-exit configuration, which runs in 0.42–0.48 s with a recall of 0.81, compared with 0.855 for the unoptimized model. The application also supports hardware acceleration through NNAPI, which dispatches inference to available GPU, NPU, or DSP accelerators. Overall, this work demonstrates that YOLO-based real-time vision systems can be effectively adapted for edge deployment on resource-constrained devices. By combining accuracy-preserving optimizations with multimodal feedback, the application enhances safety and situational awareness for a broad user base. 

The remainder of this paper is organized as follows. Section \ref{sec:related_work} reviews related work. Section \ref{sec:methodology} describes the methodology. Section \ref{sec:evaluation} presents the evaluation results. Section \ref{sec:conclusion} concludes the paper and outlines future directions.

\section{Related Work}
\label{sec:related_work}

\subsection{Mobile-phone Obstacle Detection}
The foundational challenge in mobile obstacle detection is coupling real-time detection with reliable distance estimation in a universal, on-device application. Previous studies, such as \cite{Tapu2013}, established the feasibility of on-phone pipelines for object classification, yet their motion-compensation methods resulted in a borderline frame rate and vulnerability to static obstacles. Similarly, \cite{Rahman2018} achieved high accuracy, but only by restricting the scope to indoor obstacle detection with a non-adaptive, static detection threshold. As a distance problem solution, \cite{datta2024enhanced} created a dual-purpose system combining a modern detection model (YOLOv4) with focal-length-based distance estimation. However, this work lacks real-world usability studies. This software-based monocular estimation (like \cite{Vajgl2022}'s Dist-YOLO or the YOLO-MiDaS fusion by \cite{jaimin2023estimating}) represents the most ubiquitous and accessible method, as it runs on any device. Conversely, the commercial smart-phone application \cite{Messaoudi2022} and the research of \cite{10.1108/IR-06-2018-0113} represent the high-fidelity approaches, using active sensors (LiDAR/TrueDepth) or multi-camera stereo vision for centimeter-level distance; These solutions, however, are either hardware-exclusive \cite{Messaoudi2022} or cumbersome due to rigid calibration \cite{Bertozzi1996}, sacrificing universal accessibility. The work of \cite{das2020real} offers a pragmatic middle ground, suggesting opportunistic multisensor recalibration to increase the metric fidelity of monocular estimators without continuous LiDAR reliance, leading to the clear conclusion that the most practical path is a hybrid approach that adopts the most accessible software method while opportunistically leveraging high-fidelity sensors when available.

\subsection{Real-time Obstacle Detection Across Diverse Online/Offline Environments}
Broader assistive applications extend perception beyond basic obstacle alerts, covering tasks like scene description, Optical Character Recognition (OCR), and semantic segmentation. Production examples like Microsoft \cite{microsoft_garage_seeing_ai_2025} and Google \cite{GoogleLookout2018} offer robust scene understanding with strong UX, often blending on-device and cloud processing. Similarly, the system in \cite{WeWALK2025} integrates a physical smart cane with mobile alerts and navigation. These industrial solutions, including \cite{BlindSquare2025}, prioritize usability but are often constrained by reliance on device-specific sensors or cloud services. In research, \cite{Kumar2021} presented a YOLOv3-based assistive system with Text-to-Speech feedback, while \cite{Kumar2023} demonstrated YOLOv5-based mobile recognition with efficient inference. Overall, the trend is toward hybrid solutions and integrated ecosystems, where the smartphone acts as a central hub for deep learning while partnering with specialized external sensors, as demonstrated by \cite{kasabekar2024smart} , with all solutions sharing the critical requirement of providing clear, concise audio and haptic feedback.

\subsection{CNN Architectures and End-to-End Mobile Implementations}
Some studies have demonstrated the feasibility of running advanced computer vision pipelines on constrained mobile devices, even when their focus was not obstacle detection or assistive navigation. It has been shown that ultra-low precision quantization, down to 4-bit, enables accurate inference of heavy detectors such as RetinaNet and Faster R-CNN on phones and drones \cite{Li2019}. Some research studies like \cite{Pedoeem2018} exemplified lightweight design for real-time inference on non-GPU devices. The Glimpse framework \cite{Chen2015}  combined local inference with selective offloading to reduce latency and energy use, offering principles that extend to continuous assistive pipelines. Finally, \cite{Primbetov2024} demonstrated that domain-specific lightweight detectors, such as YOLOv2-Tiny for logo recognition, can deliver smooth real-time performance on Android by narrowing the detection scope. Together, these works highlight that quantization, architectural simplification, hybrid processing strategies, and task-specific tailoring are essential tools for deploying vision models on resource-constrained devices.

\subsection{Optimizing YOLO: Compact Architectures, Quantization, Pruning, Early-Exit}
The YOLO family has become the default baseline for real-time object detection, yet running these models on smartphones requires targeted optimization. Several research studies emphasized lightweight architectures, quantization, pruning, and domain-specific tailoring. Early efforts simplified YOLO directly. Some research studies like \cite{Pedoeem2018} reduced layers and batch normalization, while \cite{Shi2019} used fire modules and dense connections for efficient deployment. Later works introduced specialized variants like \cite{Chen2024}'s YOLOv8-PSS and \cite{Gao2025}'s GCP-YOLO. Hardware-aware variants such as \cite{Yushchenko2023}, \cite{MBABYOLO2023}, and \cite{Lou2023} further optimized computation flow, while domain-specific versions like \cite{Gupta2024} showed that task-focused models outperform heavier general detectors. Furthermore, \cite{Reis2023} presents a real-time object detection system using YOLOv8 to identify flying objects like drones, birds, and helicopters. It applies transfer learning and selects the medium YOLOv8 variant to balance detection accuracy and runtime speed on real-world data.

Existing mobile obstacle detection systems show significant advances but remain limited in scope and adaptability. Prior software-based monocular methods \cite{Tapu2013, Rahman2018} often compromise between speed and accuracy or lack validation in diverse real-world environments. Sensor-dependent approaches \cite{Messaoudi2022, 10.1108/IR-06-2018-0113} achieve high precision but restrict accessibility due to proprietary hardware and calibration demands. Similarly, assistive applications \cite{Kumar2021, Kumar2023, Sanjar2024} and commercial tools \cite{microsoft_garage_seeing_ai_2025, GoogleLookout2018} rely on cloud or device-specific computation, reducing responsiveness and offline reliability. Furthermore, most prior systems detect only a narrow set of obstacle types, limiting situational awareness. In contrast, our framework detects 18 object categories in real-time and integrates adaptive sensor fusion with on-device multimodal feedback, providing a scalable, hardware-agnostic solution for robust mobile obstacle detection.

\section{Proposed Method}
\label{sec:methodology}

The proposed system consists of three primary components: (1) dataset compilation and preparation, (2) AI model optimization and architectural enhancement, and (3) Android-based implementation and deployment. Together, these stages form an integrated pipeline for real-time obstacle detection and assistive feedback on smartphones.

\subsection{Dataset Origins and Composition}
The Real-time Obstacle Detector model (YOLOv8n) was trained on a highly diverse and richly annotated dataset compiled by integrating multiple publicly available sources from Roboflow Universe \cite{people-s773n_dataset, stair-detection-project_dataset, tree-classification-qbhpo_dataset, road-users-detection-a2kmr_dataset, person-detection-9a6mk_dataset, minor_dataset-al08q_dataset, dynamic-object-detection-yioue_dataset, bin-koznr_dataset, lab1lab1_dataset, human-3uzai_dataset, tree-detect-xmqaa_dataset, crosswalk-yyiit_dataset, aml-homework-1_dataset, pothole-detection-2-wdvk0-hk1rk_dataset, road-safety-ai-guard-rails_dataset, bicycle-training-data_dataset, yolo-dataset-m1jrw_dataset, bike_detection-0rlsj_dataset, dataadd-bkykc_dataset, car-2-xdhdl_dataset, object-detection-gkkwo_dataset, bicycle-zrn08_dataset, iranian-vehicle-cjfc5_dataset, car-and-motorcycle-detection-with-kaggle-dataset_dataset, signs-sqcaz_dataset, bench-detector_dataset}. The datasets cover a broad spectrum of frequently encountered urban obstacles, including vehicles, pedestrians, street furniture, traffic elements, and environmental features. All original datasets were converted into the YOLO Darknet TXT annotation format, where each image’s labels are stored in a text file containing one line per bounding box and object ID. All coordinates are normalized between 0 and 1. Importantly, a consistent mapping of class indices was maintained across all integrated datasets, enabling seamless alignment of annotations.

From these combined sources, our final integrated dataset encompasses 18 object classes, summarized with the following instance-level statistics:
\begin{itemize}
    \item Total instances across all splits: 31,331 instances
    \item Train split: 25,110 instances
    \item Validation split: 4,493 instances
    \item Test split: 1,728 instances
\end{itemize}

Table~\ref{tab:dataset} lists the number of instances per class across all splits.

\begin{table}[t]
\centering
\caption{Overall Dataset Distribution (Total: 31,331 Instances)}
\label{tab:dataset}
\resizebox{\columnwidth}{!}{%
\begin{tabular}{l c | l c}
\hline
\textbf{Class} & \textbf{Count} & \textbf{Class} & \textbf{Count} \\ \hline
Bike & 192 & Tree & 2,075 \\
Building & 1,041 & Guard rail & 1,133 \\
Car & 5,497 & Pedestrian crosswalk & 2,493 \\
Person & 8,962 & Truck & 539 \\
Stairs & 1,007 & Bus & 893 \\
Traffic sign & 907 & Bench & 260 \\
Electrical Pole & 391 & Dustbin & 1,141 \\
Road & 682 & Dog & 728 \\
Motorcycle & 1,007 & Manhole & 2,383 \\ \hline
\end{tabular}%
}
\end{table}

\subsection{AI Model Optimization Pipeline}
\label{subsec:optimization}

The core of the system is an optimized YOLOv8n model, chosen for its balance of speed and accuracy. The full optimization pipeline includes early-exit branches, pruning, and quantization, detailed below.

\begin{figure*}[htbp]
    \centering
    \includegraphics[width=\textwidth, 
                   trim={3.5cm 8cm 2.6cm 2.5cm}, 
                   clip]{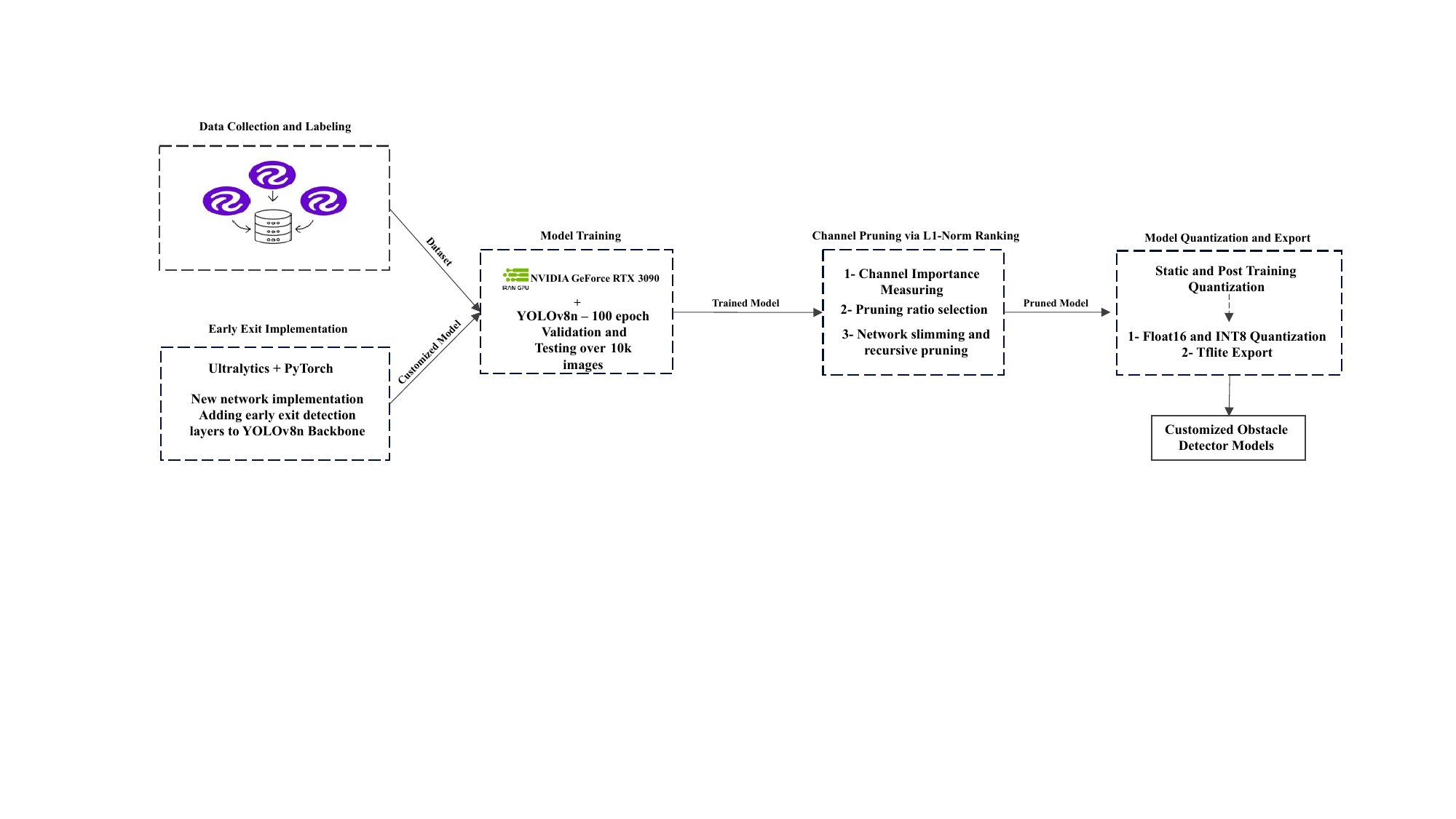}
    \caption{Overview of the YOLO architecture modifications implemented in this work, including pruning, early-exits, and quantization strategies that collectively enhance detection speed and efficiency on mobile devices.}
    \label{fig:ai_yolo_changes}
\end{figure*}

\subsubsection{Early-Exit}
Although the integration of early-exit branches into deep neural networks generally increases the training overhead and slightly slows down convergence, it provides substantial improvements during inference. The reduction in inference time is particularly significant, making the additional cost during training negligible in practical scenarios.

In this work, we incorporate early-exit points into selected C2f cells of the YOLO architecture. Our exit locations are a deliberate, task-aware and hardware-aware customization rather than fixed positions from prior work: we attach side heads immediately after C2f modules at progressively coarser strides in the backbone (after C2f in layers c2f2, c2f4, c2f6, and c2f8) and add one further branch after the first neck fusion near the c2f12 path, giving five early-exit heads in total. Analytically, C2f’s split, transform, and fuse design yields discriminative, well-conditioned feature tensors at these stages, making them suitable tap points for side heads without perturbing the main flow; moreover, the associated receptive fields already suffice for large and medium obstacles, so exiting here maximizes floating-point operations (FLOPs) savings by bypassing deeper Path Aggregation Network (PAN) and Feature Pyramid Network (FPN) aggregation when confidence is high. The final neck-side exit leverages the first multi-scale fusion to catch smaller or partially occluded instances before full-depth processing. 

When objects are confidently detected at a shallower exit, inference terminates there without propagating the input through the deeper layers, reducing latency and computational cost on low-complexity inputs.

\subsubsection{Pruning}
Deep convolutional neural networks (CNNs) such as YOLOv8 have achieved impressive results in object detection and have been applied in tasks such as real-time obstacle detection. However, their high parameter count and computational demands make direct deployment on embedded and resource-constrained platforms challenging. Model compression techniques are therefore necessary to enable real-time inference. In this work, pruning is applied to reduce model size and computation; its effect on detection accuracy is evaluated in Section~\ref{sec:evaluation}.

Pruning is the process of reducing the size and complexity of a neural network by removing less important parameters, usually weights or entire channels. Instead of lowering the precision of values to reduce computation costs (as in quantization), pruning directly eliminates parts of the network that contribute least to performance. This process lowers the number of trainable parameters, decreases memory requirements, and accelerates inference by simplifying computations. While pruning may introduce some accuracy loss, the goal is to achieve a lighter model with nearly identical detection quality.

Channel Pruning via L1-Norm Ranking:

We implement pruning using an importance-based channel selection mechanism. The procedure is outlined as follows:
\begin{itemize}
    \item Channel Importance Measurement: The importance of each convolutional channel is estimated using the L1-norm of its weights. Channels with smaller L1-norms are considered less informative.
    \item Pruning Ratio Selection: A user-defined pruning ratio $r \in [0, 1]$ specifies the proportion of channels to remove from each pruned layer. In our final model, each pruned layer retains the $50\%$ of channels with the highest L1-norm.
    \item Network Slimming: Convolutional layers are reconstructed with the surviving channels ($slim\_conv$), and corresponding Batch Normalization layers are adjusted to match ($slim\_bn$).
    \item Pruning Scope: The pruning function traverses the network hierarchy and prunes the hidden $3 \times 3$ convolution inside each C2f bottleneck. The $1 \times 1$ convolutions are excluded, and the bottleneck output convolutions keep their width because their outputs are added to residual connections; this preserves the structural integrity of the network.
\end{itemize}
Overall, pruning reduces the model from 3.01M to 2.54M parameters, a $15.7\%$ reduction (Table~\ref{tab:opt_summary}).
Structured channel pruning with L1-norm ranking reduced the size and latency of our YOLOv8 model. However, as reported in Section~\ref{sec:evaluation}, it also lowered recall substantially (from 0.855 to 0.562), so for this safety-critical application we treat pruning as a speed--recall trade-off rather than a free improvement.

\subsubsection{Quantization}
We add quantization as a final, deployment-oriented optimization to reduce model size, memory traffic, and latency. Concretely, we export two TensorFlow Lite variants of each model configuration:
\begin{itemize}
    \item An INT8 model produced by post-training dynamic-range quantization: convolution weights are stored as 8-bit integers, while activations remain in FP32 and weights are dequantized on the fly during inference.
    \item An FP16 model that stores all weights in half precision, for devices with efficient half-precision support.
\end{itemize}
Dynamic-range quantization requires no calibration dataset. In the pruned model, 97.8\% of the parameters are stored as INT8; only small tensors (fewer than 1,024 values, such as the first convolution and the biases) remain in FP32. This reduces the TFLite model size by 72\% (from 10.2\,MB to 2.8\,MB), and the FP16 export reduces it by 50\% (to 5.1\,MB).

We target the TFLite format specifically to integrate the quantized and customized variants with our TensorFlow-based detector class on Android (detailed later), allowing us to choose the best artifact per device: INT8 for CPU execution, and FP16 where GPU paths offer strong half-precision throughput.

In conclusion, quantization complements pruning and early exits by delivering smaller models and measurable speedups at deployment time without retraining.

\subsection{Android Implementation}

The Android application is presented as a user-centered on-device perception pipeline that combines optimized local AI with pragmatic system engineering to deliver low-latency obstacle awareness and accessible feedback. By executing sensing, inference, and decision logic directly on the phone, latency is reduced, user privacy is better protected, and the application remains functional without a network connection. Key design priorities included minimizing inference delay, maximizing compatibility across heterogeneous Android hardware, and ensuring reliable operation under varied real-world conditions such as changing illumination, user motion, and scene clutter.

\begin{figure}[t]
\centering
\adjustbox{trim={0 0 0 0cm}, clip}{%
    \includegraphics[
    width=\columnwidth,
    keepaspectratio]{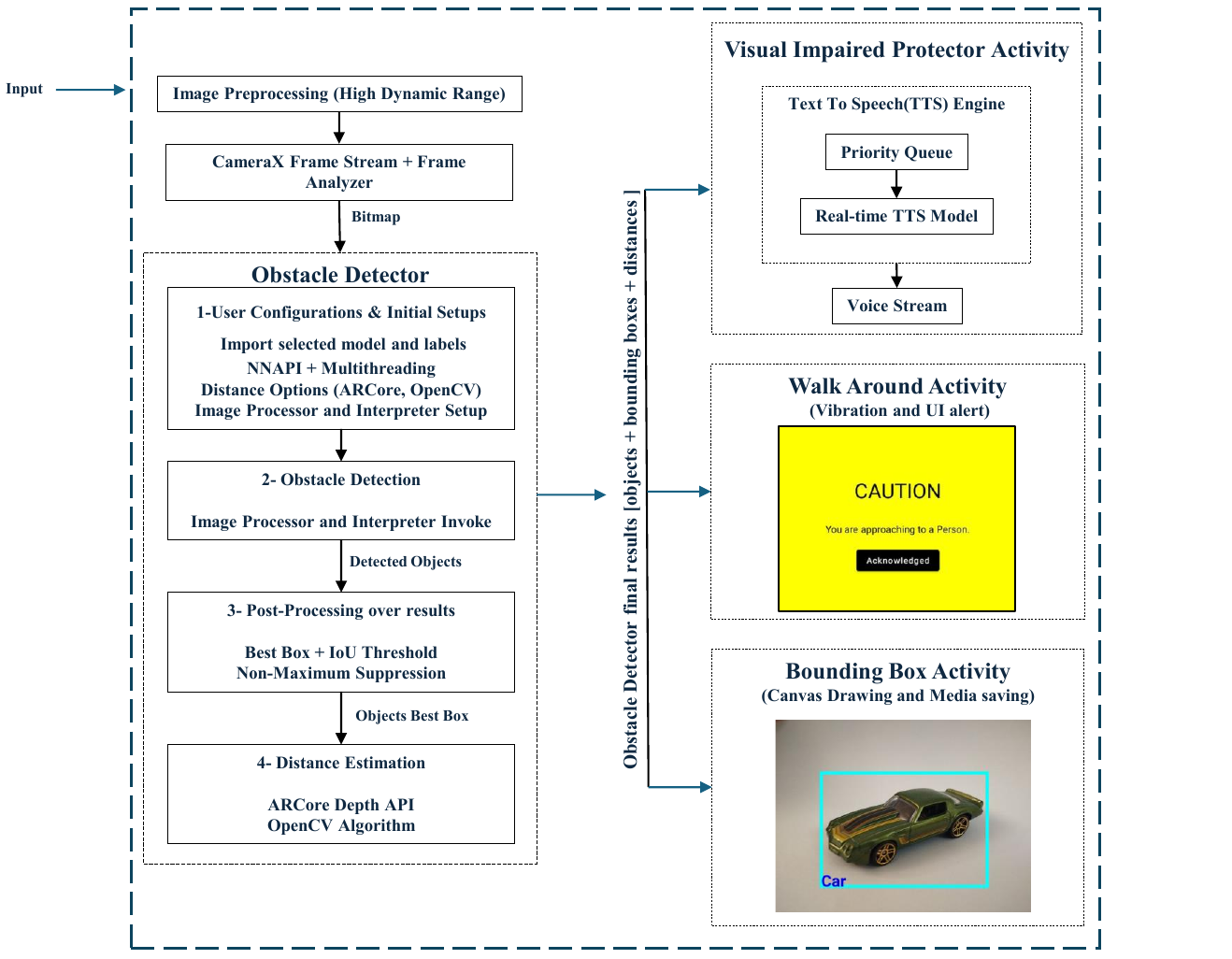}       
}
\caption{Schematic of the proposed Android application runtime pipeline, showing the sequential flow from camera acquisition and preprocessing through inference, post-processing, and multimodal output generation.}
\label{fig:structure}
\end{figure}

Conceptually, the system begins with an initial configuration and device-capability probe, proceeds to a CameraX-based frame acquisition and lightweight preprocessing pipeline (including HDR adjustment and normalization), then performs NNAPI-backed model inference followed by IoU filtering, non-maximum suppression, and priority ranking. The process is finalized by delivering multimodal outputs consisting of on-screen bounding boxes with optional media saving, graduated haptic alerts, and prioritized Text-to-Speech narration. Runtime instrumentation for frames per second (FPS) and per-frame timing and deterministic resource cleanup complete the processing flow. A schematic of the runtime pipeline is provided in Figure~\ref{fig:structure}, and the following subsections describe the implementation rationale for each stage in greater detail.

\subsubsection{App Entry, Configuration Card, and Mode Selection}
Upon first launch, the user is presented with a lightweight configuration card that enables selection of an operating mode (General Pedestrian or Visually Impaired) and configuration of runtime parameters such as distance thresholds, IoU and confidence thresholds, and preferred feedback channels (e.g., vibration, Text-to-Speech, and optional media saving). Options are also provided to enable or disable ARCore depth and to select inference execution preferences, including exploiting hardware accelerators when present. This choice was made to minimize inference latency and the energy cost per inference, thereby preserving battery life and keeping the UI responsive during continuous use.

\subsubsection{Permission, Capabilities Check, and Graceful Fallbacks}
After configuration, device capabilities are probed (camera availability, NNAPI-supported accelerators, ARCore support). Probing is performed up front to ensure a consistent and predictable runtime plan can be selected immediately: when an accelerator (Neural Processor Unit (NPU)/Graphics Processing Unit (GPU)/Digital Signal Processing (DSP)) is available, inference will be routed through NNAPI to reduce CPU load and latency; when ARCore is present, ARCore depth will be used for metric distance; otherwise, the focal-length-based distance estimator will be enabled. This capability-first approach maximizes the proportion of users who receive the best available performance on their device while ensuring functional continuity on devices with more limited hardware.

\subsubsection{Camera Acquisition and Robust Preprocessing}
Camera frames are acquired using CameraX, which was selected for its reliable lifecycle handling, consistent API across Android versions, and ease of binding to background analyzers. Early HDR adjustment and simple normalization are applied to each frame before inference to reduce lighting-induced variability. These preprocessing steps were selected because they improve detection stability in high-contrast outdoor scenes without introducing expensive computations. Applying these operations on-device was preferred so that model inputs remain consistent and cloud round-trips are avoided.

\subsubsection{Concurrency and On-device Execution}
Kotlin was chosen as the implementation language because it integrates idiomatically with Android APIs and enables concise, maintainable code. In addition, Kotlin Coroutines were employed to orchestrate the concurrent tasks of frame capture, model inference, and UI updates in a lightweight and deterministic manner. Coroutines were selected over manual thread management to reduce synchronization complexity and limit context-switch overhead on mobile CPUs. For model execution, the Android Neural Networks API (NNAPI) was employed to manage the device's processing behaviour, which can be tailored to both the user's needs and the device's capabilities. The app can be tuned for lower-latency operation on accelerator-capable phones or for broader compatibility on legacy devices, and it can dynamically adapt thresholds and output policies to accommodate individual preferences. Mode selection determines the output policy and priority logic; for example, concise haptic and visual alerts are routed to distracted pedestrians, while prioritized spoken messages are delivered through the Text-to-Speech queue for visually impaired users. By exposing these runtime parameters, device- and user-specific trade-offs are managed without code changes, reducing the need for device-by-device calibration and facilitating adoption across heterogeneous Android hardware.

\subsubsection{Inference, Post-processing and Prioritization}
Each preprocessed frame is passed to the on-device detector via an NNAPI-backed interpreter. IoU thresholding and non-maximum suppression are executed immediately to produce a final set of labeled bounding boxes. A lightweight priority queue then ranks detected objects by configurable rules such as proximity, class-criticality, and user mode, ensuring that urgent hazards such as a person or stairs within the configured safety distance are surfaced first. Prioritization was implemented to avoid information overload, a necessary safeguard for spoken guidance and to reduce false-alarm fatigue.

\subsubsection{Distance Estimation Strategy}
Spatial awareness is provided opportunistically: when ARCore is available, ARCore depth is used for higher-fidelity distance estimates (preferred for close-range safety-critical cues). On devices without ARCore, metric distance is estimated from a single RGB frame using a focal-length perspective projection method implemented with OpenCV and CameraX intrinsics. For each detected object class, a canonical real-world height $H$ (in metres) is assumed (for example, person $\approx 1.6\,\text{m}$, car $\approx 1.8\,\text{m}$). The detector outputs bounding boxes in normalized coordinates; the bounding-box height is converted to pixels ($h_{\text{px}}$), and the following perspective relation is applied:

\begin{Equation}
\begin{equation}
Z = \frac{H \cdot f_{\text{px}}}{h_{\text{px}}}
\end{equation}
\end{Equation}

Where $Z$ is the estimated distance in metres, $H$ the assumed object height, $f_{\text{px}}$ the camera focal length in pixels, and $h_{\text{px}}$ the bounding-box height in pixels. The detector's normalized center-y and normalized height are first transformed to pixel units using the model tensor/image height; the top and bottom pixel coordinates are obtained from the center and pixel-height. Focal length in pixels is computed from sensor intrinsics when available or from millimetre focal length and physical sensor size as:

\begin{Equation}
\begin{equation}
f_{\text{px}} = f_{\text{mm}} \times \frac{\text{ImageHeight}_{\text{px}}}{\text{SensorHeight}_{\text{mm}}}
\end{equation}
\end{Equation}

If intrinsic focal values ($f_{x}$, $f_{y}$) are supplied by the camera API or ARCore, they are used directly. This focal-length-based monocular method was selected because it is computationally lightweight, requires no additional hardware and provides a deterministic per-frame metric estimate suitable as a fallback on devices without depth sensors. The implementation is integrated into the CameraX analysis pipeline so that distance estimates are produced alongside the detector's bounding boxes and are consumed by the application's prioritization and alerting logic.

\subsubsection{Multimodal Outputs and Adaptive Alerting}
Detection outcomes are routed to three coordinated outputs, each tuned to the selected mode and designed to minimize cognitive load:
\paragraph{Visual overlay (Debugging)} Bounding boxes and labels are rendered for sighted users and for debugging.
\paragraph{Haptic and on-screen alerts (Walk-Around mode)} A graduated vibration scheme and an unobtrusive UI warning are emitted when obstacles are detected within mode-specific distance thresholds. Haptic feedback is prioritized because it can be perceived without visual attention and is reliable across device models. In addition, a color-coded alert system is introduced to reinforce the warnings: yellow indicates a lower-priority hazard, while red signals a high-risk or imminent danger. These choices align with widely recognized UI conventions and human cognitive models, where red is conventionally associated with critical warnings and immediate threats, and yellow communicates caution without inducing alarm.
\paragraph{Vocal guidance (Blind-Protector mode)} Detected items are enqueued by priority and delivered via Android's Text-to-Speech service. Text-to-Speech was chosen because it provides immediate, local, and private audio feedback without any network dependence and because Android's Text-to-Speech pipeline supports rapid voice parameterization and language handling. To prevent message collisions, a short-rate limiter and interrupt policy are enforced so that the spoken stream conveys the most relevant information while preserving intelligibility.

\section{Evaluation over Android Device}
\label{sec:evaluation}
\subsection{Recall and Precision Evaluation}
In order to verify the effectiveness and stability of the proposed optimization mechanisms, we conducted a series of experiments using our custom evaluation notebook. This notebook served as the official validation pipeline for all model variants, integrating the modified architecture and inference procedures described in the previous section. The evaluation quantitatively assesses how introducing early-exit branches, pruning, and quantization affects detection accuracy on the trained model.

All experiments were run with the Ultralytics YOLOv8 framework on a Tesla T4 GPU in a controlled environment to ensure deterministic execution and reproducibility. The test set comprised 1,152 annotated images containing 1,728 object instances, consistent with the split defined in the project’s data file. The notebook automatically executed the validation routine and logged the standard detection metrics.

\begin{table}[t]
\centering
\caption{Accuracy Metrics Results (Precision, Recall, mAP$_{50:95}$)}
\label{tab:accuracy_metrics}
\renewcommand{\arraystretch}{1.8}
\setlength{\tabcolsep}{4pt}
\resizebox{\columnwidth}{!}{%
\begin{tabular}{c p{3.7cm} c c c c}
\toprule
\textbf{Row} & \textbf{Model / Optimization} & \textbf{Precision} & \textbf{Recall} & \textbf{mAP$_{50:95}$}\\ 
\midrule
1 & Customized YOLO (18-classes), no optimization (FP32) & 0.843 & 0.855 & 0.727 \\
2 & Early-Exit only (Backbone+Neck, FP32) & 0.885 & 0.823 & 0.725\\
3 & Pruning only (Customized, FP32) & 0.839 & 0.562 & 0.476\\
4 & Early-Exit + Pruning (FP32) & 0.871 & 0.534 & 0.446 \\
5 & INT8 only (Customized) & 0.898 & 0.790 & 0.707 \\
6 & FP16 only (Customized) & 0.892 & 0.794 & 0.708\\
7 & INT8 + Early-Exit & 0.877 & 0.810 & 0.711 \\
8 & FP16 + Early-Exit & 0.875 & 0.807 & 0.711\\
9 & INT8 + Pruning (Customized) & 0.831 & 0.555 & 0.473\\
10 & FP16 + Pruning (Customized) & 0.839 & 0.562 & 0.474 \\
11 & INT8 + Early-Exit + Pruning & 0.873 & 0.539 & 0.446\\
12 & FP16 + Early-Exit + Pruning & 0.871 & 0.534 & 0.445 \\
\bottomrule
\end{tabular}
}
\end{table}

The resulting metrics are summarized in Table~\ref{tab:accuracy_metrics}, which reports the outputs of the YOLOv8 validation engine for each variant; on-device latency is evaluated separately in Section~\ref{subsec:runtime}.

The baseline customized model achieves strong balance, with high precision and the highest recall. Since our goal is to avoid missing any obstacles, recall is the main focus throughout this comparison. Adding early-exit layers at the backbone and neck increases precision but slightly reduces recall, while overall mAP remains nearly unchanged—indicating a trade-off between sensitivity and specificity. Applying quantization without pruning (INT8 or FP16) maintains accuracy close to the baseline. Combining quantization with early-exit recovers some recall and keeps high precision, offering a practical speed–accuracy compromise for on-device deployment.

Introducing pruning, whether alone or combined with early-exit or quantization, leads to a marked decrease in recall and mAP.
Given the goal of “don’t miss obstacles,” pruning is not suitable for this application, and configurations that preserve recall should be preferred even if slightly slower. In conclusion, for maximum recall, the baseline FP32 remains best (0.855). If runtime efficiency is needed, INT8 + EE (no pruning) offers the best recall among optimized variants ($\sim0.810$) while keeping precision high.

\subsection{Runtime and Performance Evaluation}
\label{subsec:runtime}
We evaluated multiple YOLO-family variants on a Samsung Galaxy M32 Android device. Each configuration was profiled on approximately 33\% of our internal test set. We report average per-image inference time (in milliseconds). For each model, we tested four runtime settings to isolate pre-/co-processing effects: (i) No HDR, No NNAPI, (ii) HDR only, (iii) NNAPI only, and (iv) HDR + NNAPI. EE was implemented once at the Backbone+Neck stage. The customized detector targets 18 obstacle classes.

For our evaluation, no directly comparable end-to-end system was available that could provide a balanced and fair benchmark, and most rival applications did not offer TFLite or PyTorch models suitable for execution inside our Android pipeline. This made a strict model-to-model comparison infeasible. Instead, we selected one related YOLOv8-based real-time detector \cite{Reis2023} and reproduced its selected model, YOLOv8m, within our own test environment. It appears as Row~14 of Table~\ref{tab:runtime_results}, evaluated on the same device and test images as our models. Because \cite{Reis2023} uses the larger medium variant, this comparison illustrates the cost of deploying a larger, unoptimized detector on the device rather than a like-for-like comparison of optimization methods.

\begin{table}[t]
\centering
\caption{Inference Time Comparison on Samsung Galaxy M32 (ms)}
\label{tab:runtime_results}
\renewcommand{\arraystretch}{1.8}
\setlength{\tabcolsep}{4pt}
\resizebox{\columnwidth}{!}{%
\begin{tabular}{c p{3.7cm} c c c c}
\toprule
\textbf{Row} & \textbf{Model / Optimization} & \textbf{HDR+NNAPI} & \textbf{NNAPI Only} & \textbf{HDR Only} & \textbf{None} \\
\midrule
1 & YOLOv8 base (untrained/customization-free, FP32) & \textbf{1077.44} & 1081.23 & 3205.27 & 3209.32 \\
2 & Customized YOLO (18-classes), no opt (FP32) & \textbf{613.27} & 622.61 & 993.95 & 992.19 \\
3 & Early-Exit only (Backbone+Neck, FP32) & \textbf{622.21} & 649.84 & 995.01 & 987.87 \\
4 & Pruning only (Customized, FP32) & 431.54 & \textbf{419.90} & 881.25 & 883.79 \\
5 & Early-Exit + Pruning (FP32) & \textbf{429.95} & 432.99 & 859.41 & 856.40 \\
6 & INT8 only (Customized) & \textbf{502.07} & 530.61 & 553.58 & 552.03 \\
7 & FP16 only (Customized) & 992.32 & 1256.68 & 1002.92 & \textbf{942.50} \\
8 & INT8 + Early-Exit & 430.83 & 426.82 & \textbf{419.81} & 475.16 \\
9 & FP16 + Early-Exit & 1025.11 & 1021.76 & 1035.47 & \textbf{1003.34} \\
10 & INT8 + Pruning (Customized) & 399.10 & 437.54 & 397.81 & \textbf{393.39} \\
11 & FP16 + Pruning (Customized) & 856.88 & 858.91 & 855.51 & \textbf{854.86} \\
12 & INT8 + Early-Exit + Pruning & \textbf{413.81} & 427.14 & \textbf{395.18} & 397.84 \\
13 & FP16 + Early-Exit + Pruning & 880.08 & 880.20 & 881.73 & \textbf{859.32} \\
\midrule
14 & Reis et al. \cite{Reis2023} (YOLOv8m)$^{\dagger}$ & 2346.04 & 2724.87 & 8838.08 & 8853.30 \\
\bottomrule
\multicolumn{6}{l}{\footnotesize $^{\dagger}$Reproduced on the same device and test images as Rows 1--13.}
\end{tabular}%
}
\end{table}

The off-the-shelf YOLOv8 model (Row~1) is significantly slower on the M32 device ($\approx$3.2\,s without NNAPI). Customizing the model for the 18-class domain (Row~2) reduces latency to $\approx$613–623\,ms under NNAPI, 1.6× faster than its own CPU-only inference ($\approx$992\,ms) and 5.2× faster than the unmodified base model on the CPU ($\approx$3.2\,s).

Early-Exit alone at FP32 (Row~3) did not outperform the customized FP32 model without optimization under NNAPI (Row~2). Its latency benefit becomes evident when combined with INT8: INT8 + Early-Exit (Row~8) is 14–24\% faster than INT8 alone (Row~6) across all four runtime settings. Pruning the customized FP32 model (Row~4) brought latency from $\sim992$\,ms (CPU) down to $\sim881$--$884$\,ms and to $\sim420$--$432$\,ms with NNAPI. Adding EE to the pruned FP32 model (Row 5) gives a very similar latency envelope, suggesting limited extra gain from EE when the model is already pruned at FP32.

Quantization to INT8 (Row~6) consistently outperforms FP16 (Row~7), while INT8 + Pruning (Row~10) achieves the best latency ($\approx$393–437\,ms). FP16 variants remain slower on this device. NNAPI reduces FP32 latency by 34–66\% (Rows 1--5). HDR preprocessing has no consistent effect: for most configurations the HDR and non-HDR settings differ by less than 10\%, although larger differences occur in a few cases (e.g., Rows 7 and 8), which likely reflect measurement variance.

The reproduced YOLOv8m model of \cite{Reis2023} (Row~14) requires $2346$--$2725$\,ms under NNAPI and $\approx 8.8$\,s on the CPU. All of our customized variants are faster in every runtime setting; for example, the recommended INT8 + Early-Exit model (Row~8) is 5.4–21× faster, showing that a smaller, optimized detector is essential for on-device use on a mid-range phone such as the Galaxy~M32.

For pedestrian obstacle detection, the real-time requirement is set by walking speed rather than video frame rate. At an average walking speed of about 5\,km/h ($\approx$1.4\,m/s), a user moves approximately 0.7\,m during a 0.5\,s inference cycle; we therefore adopt 0.5\,s per frame as the real-time budget for this application. With this budget, an obstacle 5\,m ahead is re-evaluated about seven times before the user reaches it. On the Galaxy~M32, INT8 + Early-Exit (Row~8), INT8 + Pruning (Row~10), and INT8 + Early-Exit + Pruning (Row~12) meet this budget in all four runtime settings (0.39–0.48\,s), as do the pruned FP32 models under NNAPI (Rows 4--5). Moving obstacles such as bicycles and vehicles shorten this margin because the relevant quantity is the closing speed: a bicycle at 15\,km/h approaching a walking user closes the gap by about 2.8\,m per cycle, which motivates further latency reduction in future work.

\begin{figure}[t]
\centering
\adjustbox{trim={0 0 0 0.3cm}, clip}{%
    \includegraphics[width=\columnwidth]{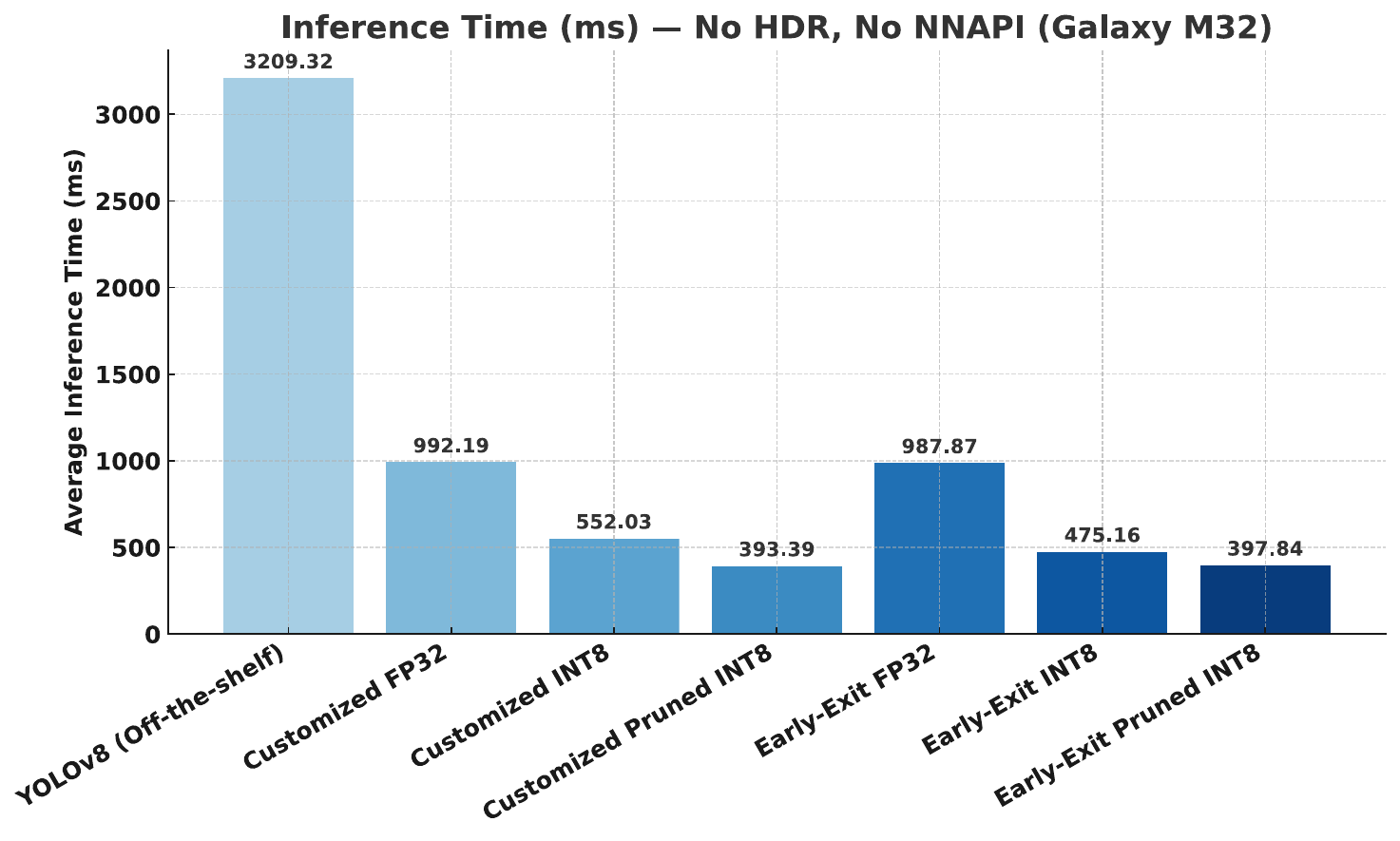}             
}
\caption{Inference performance of YOLOv8-based configurations on Samsung Galaxy M32 without hardware acceleration or HDR preprocessing. Lower values indicate faster performance.}
\label{fig:eval_bar}
\end{figure}

Figure~\ref{fig:eval_bar} compares the inference performance of seven representative YOLOv8-based configurations when executed on the Samsung Galaxy M32 without any hardware acceleration or HDR preprocessing. The off-the-shelf YOLOv8 model demonstrates the slowest inference time ($\approx3.2$ s per frame), confirming that the unmodified architecture is not suitable for real-time operation on mid-range mobile devices, while quantization and pruning significantly improve speed. The fastest configurations, INT8 + Pruning and INT8 + Early-Exit + Pruning, achieve 0.39–0.40 s per frame in CPU-only mode. These results confirm that quantization and pruning have a greater effect on inference latency than early exit alone in mobile CPU environments.

\subsection{Model Optimization Summary}

Table~\ref{tab:opt_summary} summarizes the effect of each optimization on model size. In the INT8 export, 97.8\% of the parameters are stored as 8-bit integers, reducing the TFLite model from 10.2\,MB to 2.8\,MB. The INT8 model stays close to the baseline in accuracy (recall 0.790 vs.\ 0.855, mAP$_{50:95}$ 0.707 vs.\ 0.727; Table~\ref{tab:accuracy_metrics}) while reducing CPU inference time from $\approx$992\,ms to $\approx$552\,ms. The FP16 export stores all weights in half precision, halving the model size to 5.1\,MB with similar accuracy (recall 0.794), but it was not faster on this device.

Pruning removes 472,944 parameters (15.7\%), reducing the model from 3.01M to 2.54M parameters. This lowers the number of multiply--accumulate operations and contributes to the observed speedup, but it also reduces recall to 0.562, which is why we do not recommend it for this application. In contrast, the early-exit heads add 2.02M parameters (+67.0\%): early exit reduces the computation performed per frame, not the model size. Combining INT8 quantization with early exit yields 0.42–0.48\,s per frame with a recall of 0.810, which we consider the best trade-off between speed and recall for obstacle detection.

\begin{table}[!t]
\centering
\caption{Model Optimization Summary}
\label{tab:opt_summary}
\renewcommand{\arraystretch}{1.4}
\setlength{\tabcolsep}{4pt}
\resizebox{\columnwidth}{!}{%
\begin{tabular}{l r r l}
\toprule
\textbf{Model Variant} & \textbf{Parameters} & \textbf{Change} & \textbf{Weight precision} \\
\midrule
Customized YOLOv8n (baseline) & 3,014,358 & --- & FP32 \\
Early-Exit (+5 exit heads)    & 5,033,374 & +67.0\% & FP32 \\
Pruned                        & 2,541,414 & $-$15.7\% & FP32 \\
Pruned, FP16 (TFLite, 5.1\,MB) & 2,587,014$^{*}$ & --- & 100\% FP16 \\
Pruned, INT8 (TFLite, 2.8\,MB) & 2,587,014$^{*}$ & --- & 97.8\% INT8 \\
\bottomrule
\multicolumn{4}{l}{\footnotesize $^{*}$Includes export constants (anchor grid, strides). FP32 TFLite: 10.2\,MB.}
\end{tabular}%
}
\end{table}

\section{Conclusion and Future Work}
\label{sec:conclusion}

This study presented a real-time, smartphone-based obstacle detection and distance estimation system designed to enhance safety and mobility for visually impaired and distracted pedestrians. By integrating an optimized YOLO-based detection network with ARCore-supported or monocular distance estimation, the proposed application delivers high-accuracy obstacle awareness without relying on external sensors or cloud connectivity. Model optimization techniques, including quantization, early-exit branches, and structured pruning, proved essential in achieving practical performance on mobile hardware. The recommended INT8 + early-exit configuration reduced inference latency by more than 80\% compared to the off-the-shelf YOLOv8n model (from $\approx$3.2\,s to 0.42–0.48\,s per frame, i.e., about 2.1–2.4 frames per second on a mid-range Samsung Galaxy M32, within the 0.5\,s real-time budget for walking pedestrians) while retaining a recall of 0.81. Pruning reduced latency slightly further but lowered recall substantially, making it unsuitable for this safety-critical task. The Android implementation prioritizes accessibility, inclusivity, and low latency through on-device processing, multimodal feedback, and hardware-aware inference acceleration via NNAPI. These results support the feasibility of turning everyday smartphones into low-cost assistive tools for enhanced pedestrian safety. Future work will include a user study with visually impaired participants and distracted pedestrians to evaluate how the Text-to-Speech and haptic feedback affect awareness, reaction time, and cognitive load; an evaluation across lighting and scene conditions; latency measurements on additional devices and accelerator backends; and extending detection coverage to smaller and low-contrast obstacles. Additional directions include expanding the training dataset for greater environmental diversity, integrating adaptive scene understanding modules (such as semantic segmentation or depth fusion), and leveraging federated or on-device continual learning to personalize performance for individual users. Long-term, this approach can evolve into a unified assistive perception framework that dynamically adapts to user context and device capabilities, further advancing inclusive and intelligent urban mobility.

\bibliographystyle{IEEEtran}
\bibliography{references}

@misc{WHO2019,
  author       = {World Health Organization},
  title        = {World Report on Vision},
  year         = {2019},
  publisher    = {World Health Organization},
  url          = {https://www.who.int/publications/i/item/world-report-on-vision},
  note         = {Accessed Oct. 26, 2025}
}

@article{VisionLoss2021,
  author  = {{GBD 2019 Blindness and Vision Impairment Collaborators} and {Vision Loss Expert Group of the Global Burden of Disease Study}},
  year    = {2021},
  title   = {Trends in prevalence of blindness and distance and near vision impairment over 30 years: An analysis for the {Global Burden of Disease Study}},
  journal = {The Lancet Global Health},
  volume  = {9},
  number  = {2},
  pages   = {e130--e143},
  doi     = {10.1016/S2214-109X(20)30425-3}
}

@misc{WHO2023,
  author       = {World Health Organization},
  title        = {World Health Organization Vision Impairment and Blindness Fact Sheet},
  year         = {2023},
  url          = {https://www.who.int/news-room/fact-sheets/detail/blindness-and-visual-impairment},
  note         = {Accessed Oct. 26, 2025}
}

@article{Stavrinos2018,
  author  = {Stavrinos, Despina and Pope, Christopher N. and Shen, Jie and Schwebel, David C.},
  year    = {2018},
  title   = {Distracted walking, bicycling, and driving: Systematic review and meta-analysis of mobile technology and youth crash risk},
  journal = {Child Development},
  volume  = {89},
  number  = {1},
  pages   = {118--128},
  doi     = {10.1111/cdev.12827}
}

@article{Pelicioni2023,
  author  = {Pelicioni, Patricia H. S. and Chan, Lily L. Y. and Shi, Shuo and Wong, Keith and Kark, Lucy and Okubo, Yosuke and Brodie, Matthew A.},
  year    = {2023},
  title   = {Impact of mobile phone use on accidental falls risk in young adult pedestrians},
  journal = {Heliyon},
  volume  = {9},
  number  = {8},
  pages   = {e18366},
  doi     = {10.1016/j.heliyon.2023.e18366}
}

@inproceedings{Tapu2013,
  author       = {Tapu, Ruxandra and Mocanu, Bogdan and Bursuc,Andrei and Zaharia, Titus},
  title        = {A Smartphone-Based Obstacle Detection and Classification System for Assisting Visually Impaired People},
  booktitle      = {2013 IEEE International Conference on Computer Vision Workshops (ICCVW)},
  pages        = {444-451},
  year         = {2013},
  doi          = {10.1109/ICCVW.2013.65}
}

@article{Messaoudi2022,
author = {Messaoudi, Mohamed Dhiaeddine and Menelas, Bob and Mcheick, Hamid},
year = {2022},
eid = {7888},
title = {Review of Navigation Assistive Tools and Technologies for the Visually Impaired},
volume = {22},
journal = {Sensors},
doi = {10.3390/s22207888},
number ={20}
}

@article{Kumar2023,
  author  = {Kumar, Harshit and Paul, Anna and Anjana, V. P. and Babu, Anandakrishnan S. and Gopakumar, V. T.},
  title   = {Empowering the visually impaired: A {YOLO-v5 CNN-based} android app for currency recognition and object detection},
  journal = {World Journal of Advanced Research and Reviews},
  year    = {2023},
  volume  = {19},
  number  = {2},
  pages   = {1459--1466},
  doi     = {10.30574/wjarr.2023.19.2.1625}
}

@article{Primbetov2024,
  author  = {Primbetov, Abbaz and Saidova, Fazilat and Yembergenova, Ulmira and Primbetov, Aziz},
  title   = {REAL TIME LOGO RECOGNITION USING YOLO ON ANDROID},
  journal = {Modern Science and Research},
  year    = {2024},
  volume  = {3},
  number  = {1},
  pages   = {1--5}
}

@inproceedings{datta2024enhanced,
  title={Enhanced Environmental Perception for Visually Impaired: A Real-time Object Detection and Distance Estimation Approach},
  author={Datta, Abhishikta and Sarkar, Ayon and Bhattashali, Abhiraj and Meruva, Akash and Manikandan, VM},
  booktitle={2024 IEEE International Conference on Interdisciplinary Approaches in Technology and Management for Social Innovation (IATMSI)},
  volume={2},
  pages={1--6},
  year={2024},
  organization={IEEE}
}

@article{Vajgl2022,
  author  = {Vajgl, Marek and Hurtik, Petr and Nejezchleba, Tom{\'a}{\v{s}}},
  title   = {Dist-YOLO: Fast Object Detection with Distance Estimation},
  journal = {Applied Sciences},
  year    = {2022},
  volume  = {12},
  number  = {3},
  pages   = {1354},
  doi     = {10.3390/app12031354}
}

@online{jaimin2023estimating,
  author       = {Jaimin-K},
  title        = {Estimating Pedestrian Proximity Using MiDaS and YOLOv7},
  year         = {2023},
  publisher = {Medium},
  url          = {https://medium.com/@jaimin-k/estimating-pedestrian-proximity-using-midas-and-yolov7-14b5a08b8740},
  note         = {Accessed Oct. 26, 2025}
}

@article{10.1108/IR-06-2018-0113,
    author = {Zhang, Xinyu and Zhou, Mo and Qiu, Peng and Huang, Yi and Li, Jun},
    title = {Radar and vision fusion for the real-time obstacle detection and identification},
    journal = {Industrial Robot: the international journal of robotics research and application},
    volume = {46},
    number = {3},
    pages = {391-395},
    year = {2019},
    month = {06},
    doi = {10.1108/IR-06-2018-0113},
    url = {https://doi.org/10.1108/IR-06-2018-0113},
    eprint = {https://www.emerald.com/ir/article-pdf/46/3/391/2725014/ir-06-2018-0113.pdf},
}

@inproceedings{Bertozzi1996,
  author    = {Bertozzi, Massimo and Broggi, Alberto and Fascioli, Alessandra},
  title     = {Real-Time Obstacle Detection using Stereo Vision},
  booktitle = {Proceedings of the VIII European Signal Processing Conference (EUSIPCO 1996)},
  year      = {1996}
}

@inproceedings{das2020real,
  title={Real time distance and mobility detection of obstacles using a smart multisensor framework for visually impaired people},
  author={Das, Dola and Saha, Prottoy and Tabassum, Nawshin},
  booktitle={2020 IEEE Region 10 Symposium (TENSYMP)},
  pages={590--593},
  year={2020},
  organization={IEEE}
}

@misc{microsoft_garage_seeing_ai_2025,
  author       = {{Microsoft Garage}},
  title        = {{Seeing AI}},
  year         = {2025},
  publisher    = {Microsoft},
  url          = {https://www.microsoft.com/en-us/garage/wall-of-fame/seeing-ai},
  note         = {Accessed Oct. 26, 2025}
}

@misc{GoogleLookout2018,
  author = {Google},
  title  = {Lookout: an app to help blind and visually impaired people learn about the world around them},
  year   = {2018},
  url    = {https://blog.google/outreach-initiatives/accessibility/lookout-app-help-blind-and-visually-impaired-people-learn-about-their-surroundings},
  note   = {Accessed Oct. 26, 2025}
}

@misc{WeWALK2025,
  author       = {WeWALK},
  title        = {WeWALK Smart Cane},
  year         = {2025},
  url          = {https://wewalk.io/}
}

@misc{BlindSquare2025,
  author       = {BlindSquare},
  title        = {BlindSquare App},
  year         = {2025},
  url          = {https://www.blindsquare.com/}
}

@inbook{Kumar2021,
author = {Kumar, D. and Thakkar, Hiren Kumar and Merugu, Suresh and Gunjan, Vinit and Gupta, Suneet},
year = {2022},
month = {01},
pages = {1631-1642},
title = {Object Detection System for Visually Impaired Persons Using Smartphone},
isbn = {978-981-16-3689-9},
doi = {10.1007/978-981-16-3690-5_154},
publisher = {ICDSMLA 2020: Proceedings of the 2nd International Conference on Data Science, Machine Learning and Applications}
}

@article{Sanjar2024,
  author       = {Sanjar, K. and Bang, S. and Ryue, S. and Jung, H.},
  title        = {Deep Learning-Based Mobility Assistance for the Blind},
  journal      = {Sensors},
  volume       = {24},
  number       = {2},
  year         = {2024},
  doi          = {10.3390/s24020541}
}

@inproceedings{Li2019,
  author    = {Li, Rundong and Wang, Yan and Liang, Feng and Qin, Hongwei and Yan, Junjie and Fan, Rui},
  title     = {Fully Quantized Network for Object Detection},
  booktitle = {Proceedings of the IEEE/CVF Conference on Computer Vision and Pattern Recognition (CVPR)},
  year      = {2019},
  pages     = {2810--2819},
  doi       = {10.1109/CVPR.2019.00292}
}

@inproceedings{Pedoeem2018,
  author    = {Huang, Rachel and Pedoeem, Jonathan and Chen, Cuixian},
  title     = {{YOLO-LITE}: A Real-Time Object Detection Algorithm Optimized for Non-GPU Computers},
  booktitle = {2018 IEEE International Conference on Big Data (Big Data)},
  year      = {2018},
  pages     = {2503--2510},
  doi       = {10.1109/BigData.2018.8621865}
}

@inproceedings{Chen2015,
  author    = {Chen, Tiffany Yu-Han and Ravindranath, Lenin and Deng, Shuo and Bahl, Paramvir and Balakrishnan, Hari},
  title     = {Glimpse: Continuous, Real-Time Object Recognition on Mobile Devices},
  booktitle = {Proceedings of the 13th ACM Conference on Embedded Networked Sensor Systems (SenSys '15)},
  year      = {2015},
  pages     = {155--168},
  doi       = {10.1145/2809695.2809711}
}

@article{Shi2019,
  author  = {Fang, Wei and Wang, Lin and Ren, Peiming},
  title   = {Tinier-{YOLO}: A Real-Time Object Detection Method for Constrained Environments},
  journal = {IEEE Access},
  volume  = {8},
  pages   = {1935--1944},
  year    = {2020},
  doi     = {10.1109/ACCESS.2019.2961959}
}

@article{Chen2024,
  author  = {Chen, Zhijian and Fang, Yijun and Yin, Jianjun and Lv, Shiyu and Sheikh Muhammad, Farhan and Liu, Lu},
  title   = {A Novel Lightweight {YOLOv8-PSS} Model for Obstacle Detection on the Path of Unmanned Agricultural Vehicles},
  journal = {Frontiers in Plant Science},
  volume  = {15},
  pages   = {1509746},
  year    = {2024},
  doi     = {10.3389/fpls.2024.1509746},
  url     = {https://www.frontiersin.org/journals/plant-science/articles/10.3389/fpls.2024.1509746/full}
}

@article{Gao2025,
  author  = {Gao, Yu and Li, Zhanying and Zhang, Kangye and Kong, Lingyan},
  title   = {{GCP-YOLO}: A Lightweight Underwater Object Detection Model Based on {YOLOv7}},
  journal = {Journal of Real-Time Image Processing},
  volume  = {22},
  number  = {1},
  pages   = {3},
  year    = {2025},
  doi     = {10.1007/s11554-024-01586-7}
}

@article{Yushchenko2023,
  author  = {Wong, Alexander and Famouri, Mahmoud and Shafiee, Mohammad Javad and Li, Francis and Chwyl, Brendan and Chung, Jonathan},
  title   = {{YOLO Nano}: a Highly Compact You Only Look Once Convolutional Neural Network for Object Detection},
  journal = {arXiv preprint arXiv:1910.01271},
  year    = {2019},
  url     = {https://arxiv.org/abs/1910.01271}
}

@article{MBABYOLO2023,
  author  = {Zhang, Jun and Meng, Yizhen and Yu, Xiaohui and Bi, Hongjing and Chen, Zhipeng and Li, Huafeng and Yang, Runtao and Tian, Jingjun},
  title   = {{MBAB-YOLO}: A Modified Lightweight Architecture for Real-Time Small Target Detection},
  journal = {IEEE Access},
  year    = {2023},
  volume  = {11},
  pages   = {78384--78401},
  doi     = {10.1109/ACCESS.2023.3286031}
}

@article{Lou2023,
  author  = {Lou, Haitong and Duan, Xuehu and Guo, Junmei and Liu, Haiying and Gu, Jason and Bi, Lingyun and Chen, Haonan},
  title   = {{DC-YOLOv8}: Small-Size Object Detection Algorithm Based on Camera Sensor},
  journal = {Electronics},
  volume  = {12},
  number  = {10},
  pages   = {2323},
  year    = {2023},
  doi     = {10.3390/electronics12102323},
  url     = {https://www.mdpi.com/2079-9292/12/10/2323}
}

@article{Gupta2024,
  author  = {Yu, Yue and Zhou, Qi and Wang, Hao and Lv, Ke and Zhang, Lijuan and Li, Jian and Li, Dongming},
  title   = {{LP-YOLO}: A Lightweight Object Detection Network Regarding Insect Pests for Mobile Terminal Devices Based on Improved {YOLOv8}},
  journal = {Agriculture},
  volume  = {14},
  number  = {8},
  pages   = {1420},
  year    = {2024},
  doi     = {10.3390/agriculture14081420},
  url     = {https://www.mdpi.com/2077-0472/14/8/1420}
}

@misc{
                            people-s773n_dataset,
                            title = { People Dataset },
                            type = { Open Source Dataset },
                            author = { Test },
                            howpublished = { \url{ https://universe.roboflow.com/test-yinss/people-s773n } },
                            url = { https://universe.roboflow.com/test-yinss/people-s773n },
                            journal = { Roboflow Universe },
                            publisher = { Roboflow },
                            year = { 2024 },
                            month = { jun },
                            note = { visited on 2025-12-30 },
                            }

@misc{
                            stair-detection-project_dataset,
                            title = { stair detection project Dataset },
                            type = { Open Source Dataset },
                            author = { Nahid Hasan },
                            howpublished = { \url{ https://universe.roboflow.com/nahid-hasan-ong0f/stair-detection-project } },
                            url = { https://universe.roboflow.com/nahid-hasan-ong0f/stair-detection-project },
                            journal = { Roboflow Universe },
                            publisher = { Roboflow },
                            year = { 2024 },
                            month = { jul },
                            note = { visited on 2025-12-30 },
                            }

@misc{
                            tree-classification-qbhpo_dataset,
                            title = { Tree classification Dataset },
                            type = { Open Source Dataset },
                            author = { Zlata },
                            howpublished = { \url{ https://universe.roboflow.com/zlata-ccrmf/tree-classification-qbhpo } },
                            url = { https://universe.roboflow.com/zlata-ccrmf/tree-classification-qbhpo },
                            journal = { Roboflow Universe },
                            publisher = { Roboflow },
                            year = { 2024 },
                            month = { jun },
                            note = { visited on 2025-12-30 },
                            }

@misc{
                            road-users-detection-a2kmr_dataset,
                            title = { Road users detection Dataset },
                            type = { Open Source Dataset },
                            author = { Yolo Project },
                            howpublished = { \url{ https://universe.roboflow.com/yolo-project-4om8f/road-users-detection-a2kmr } },
                            url = { https://universe.roboflow.com/yolo-project-4om8f/road-users-detection-a2kmr },
                            journal = { Roboflow Universe },
                            publisher = { Roboflow },
                            year = { 2024 },
                            month = { jan },
                            note = { visited on 2025-12-30 },
                            }

@misc{
                            person-detection-9a6mk_dataset,
                            title = { Person detection Dataset },
                            type = { Open Source Dataset },
                            author = { Titulacin },
                            howpublished = { \url{ https://universe.roboflow.com/titulacin/person-detection-9a6mk } },
                            url = { https://universe.roboflow.com/titulacin/person-detection-9a6mk },
                            journal = { Roboflow Universe },
                            publisher = { Roboflow },
                            year = { 2025 },
                            month = { dec },
                            note = { visited on 2025-12-30 },
                            }

@misc{
                            minor_dataset-al08q_dataset,
                            title = { Minor\_dataset Dataset },
                            type = { Open Source Dataset },
                            author = { Test tensorflow },
                            howpublished = { \url{ https://universe.roboflow.com/test-tensorflow/minor\_dataset-al08q } },
                            url = { https://universe.roboflow.com/test-tensorflow/minor\_dataset-al08q },
                            journal = { Roboflow Universe },
                            publisher = { Roboflow },
                            year = { 2024 },
                            month = { oct },
                            note = { visited on 2025-12-30 },
                            }

@misc{
                            dynamic-object-detection-yioue_dataset,
                            title = { Dynamic Object Detection Dataset },
                            type = { Open Source Dataset },
                            author = { custom object detection },
                            howpublished = { \url{ https://universe.roboflow.com/custom-object-detection-9wvfy/dynamic-object-detection-yioue } },
                            url = { https://universe.roboflow.com/custom-object-detection-9wvfy/dynamic-object-detection-yioue },
                            journal = { Roboflow Universe },
                            publisher = { Roboflow },
                            year = { 2023 },
                            month = { sep },
                            note = { visited on 2025-12-30 },
                            }

@misc{
                            bin-koznr_dataset,
                            title = { BIN Dataset },
                            type = { Open Source Dataset },
                            author = { Madhu },
                            howpublished = { \url{ https://universe.roboflow.com/madhu-3e26l/bin-koznr } },
                            url = { https://universe.roboflow.com/madhu-3e26l/bin-koznr },
                            journal = { Roboflow Universe },
                            publisher = { Roboflow },
                            year = { 2023 },
                            month = { apr },
                            note = { visited on 2025-12-30 },
                            }

@misc{
                            lab1lab1_dataset,
                            title = { lab1lab1 Dataset },
                            type = { Open Source Dataset },
                            author = { datasets },
                            howpublished = { \url{ https://universe.roboflow.com/datasets-bcj4x/lab1lab1 } },
                            url = { https://universe.roboflow.com/datasets-bcj4x/lab1lab1 },
                            journal = { Roboflow Universe },
                            publisher = { Roboflow },
                            year = { 2024 },
                            month = { mar },
                            note = { visited on 2025-12-30 },
                            }

@misc{
                            human-3uzai_dataset,
                            title = { human Dataset },
                            type = { Open Source Dataset },
                            author = { 234 432 },
                            howpublished = { \url{ https://universe.roboflow.com/234-432-om4di/human-3uzai } },
                            url = { https://universe.roboflow.com/234-432-om4di/human-3uzai },
                            journal = { Roboflow Universe },
                            publisher = { Roboflow },
                            year = { 2023 },
                            month = { dec },
                            note = { visited on 2025-12-30 },
                            }

@misc{
                            tree-detect-xmqaa_dataset,
                            title = { Tree detect Dataset },
                            type = { Open Source Dataset },
                            author = { New Workspace },
                            howpublished = { \url{ https://universe.roboflow.com/new-workspace-bpw4i/tree-detect-xmqaa } },
                            url = { https://universe.roboflow.com/new-workspace-bpw4i/tree-detect-xmqaa },
                            journal = { Roboflow Universe },
                            publisher = { Roboflow },
                            year = { 2022 },
                            month = { feb },
                            note = { visited on 2025-12-30 },
                            }

@misc{
                            crosswalk-yyiit_dataset,
                            title = { crosswalk Dataset },
                            type = { Open Source Dataset },
                            author = { koki },
                            howpublished = { \url{ https://universe.roboflow.com/koki-jnzs9/crosswalk-yyiit } },
                            url = { https://universe.roboflow.com/koki-jnzs9/crosswalk-yyiit },
                            journal = { Roboflow Universe },
                            publisher = { Roboflow },
                            year = { 2024 },
                            month = { sep },
                            note = { visited on 2025-12-30 },
                            }

@misc{
                            aml-homework-1_dataset,
                            title = { AML Homework 1 Dataset },
                            type = { Open Source Dataset },
                            author = { AML },
                            howpublished = { \url{ https://universe.roboflow.com/aml-n0koc/aml-homework-1 } },
                            url = { https://universe.roboflow.com/aml-n0koc/aml-homework-1 },
                            journal = { Roboflow Universe },
                            publisher = { Roboflow },
                            year = { 2023 },
                            month = { aug },
                            note = { visited on 2025-12-30 },
                            }

@misc{
                            pothole-detection-2-wdvk0-hk1rk_dataset,
                            title = { pothole detection 2 Dataset },
                            type = { Open Source Dataset },
                            author = { myprojects },
                            howpublished = { \url{ https://universe.roboflow.com/myprojects-gpid1/pothole-detection-2-wdvk0-hk1rk } },
                            url = { https://universe.roboflow.com/myprojects-gpid1/pothole-detection-2-wdvk0-hk1rk },
                            journal = { Roboflow Universe },
                            publisher = { Roboflow },
                            year = { 2024 },
                            month = { sep },
                            note = { visited on 2025-12-30 },
                            }

@misc{
                            road-safety-ai-guard-rails_dataset,
                            title = { Road Safety AI - Guard Rails Dataset },
                            type = { Open Source Dataset },
                            author = { Guard Rails },
                            howpublished = { \url{ https://universe.roboflow.com/guard-rails/road-safety-ai-guard-rails } },
                            url = { https://universe.roboflow.com/guard-rails/road-safety-ai-guard-rails },
                            journal = { Roboflow Universe },
                            publisher = { Roboflow },
                            year = { 2024 },
                            month = { dec },
                            note = { visited on 2025-12-30 },
                            }

@misc{
                            bicycle-training-data_dataset,
                            title = { Bicycle Training Data Dataset },
                            type = { Open Source Dataset },
                            author = { YOLOv5 },
                            howpublished = { \url{ https://universe.roboflow.com/yolov5-sbzvs/bicycle-training-data } },
                            url = { https://universe.roboflow.com/yolov5-sbzvs/bicycle-training-data },
                            journal = { Roboflow Universe },
                            publisher = { Roboflow },
                            year = { 2022 },
                            month = { aug },
                            note = { visited on 2025-12-30 },
                            }

@misc{
                            yolo-dataset-m1jrw_dataset,
                            title = { YOLO-Dataset Dataset },
                            type = { Open Source Dataset },
                            author = { FaisalDataConverter },
                            howpublished = { \url{ https://universe.roboflow.com/faisaldataconverter/yolo-dataset-m1jrw } },
                            url = { https://universe.roboflow.com/faisaldataconverter/yolo-dataset-m1jrw },
                            journal = { Roboflow Universe },
                            publisher = { Roboflow },
                            year = { 2022 },
                            month = { aug },
                            note = { visited on 2025-12-30 },
                            }

@misc{
                            bike_detection-0rlsj_dataset,
                            title = { bike\_detection Dataset },
                            type = { Open Source Dataset },
                            author = { Deo Armanta },
                            howpublished = { \url{ https://universe.roboflow.com/deo-armanta-aedas/bike_detection-0rlsj } },
                            url = { https://universe.roboflow.com/deo-armanta-aedas/bike\_detection-0rlsj },
                            journal = { Roboflow Universe },
                            publisher = { Roboflow },
                            year = { 2022 },
                            month = { jul },
                            note = { visited on 2025-12-30 },
                            }

@misc{
                            dataadd-bkykc_dataset,
                            title = { dataadd Dataset },
                            type = { Open Source Dataset },
                            author = { data },
                            howpublished = { \url{ https://universe.roboflow.com/data-ksqzc/dataadd-bkykc } },
                            url = { https://universe.roboflow.com/data-ksqzc/dataadd-bkykc },
                            journal = { Roboflow Universe },
                            publisher = { Roboflow },
                            year = { 2024 },
                            month = { aug },
                            note = { visited on 2025-12-30 },
                            }

@misc{
                            car-2-xdhdl_dataset,
                            title = { Car 2 Dataset },
                            type = { Open Source Dataset },
                            author = { FF },
                            howpublished = { \url{ https://universe.roboflow.com/ff-j3qir/car-2-xdhdl } },
                            url = { https://universe.roboflow.com/ff-j3qir/car-2-xdhdl },
                            journal = { Roboflow Universe },
                            publisher = { Roboflow },
                            year = { 2022 },
                            month = { nov },
                            note = { visited on 2025-12-30 },
                            }

@misc{
                            object-detection-gkkwo_dataset,
                            title = { object detection Dataset },
                            type = { Open Source Dataset },
                            author = { object detection },
                            howpublished = { \url{ https://universe.roboflow.com/object-detection-jlzuw/object-detection-gkkwo } },
                            url = { https://universe.roboflow.com/object-detection-jlzuw/object-detection-gkkwo },
                            journal = { Roboflow Universe },
                            publisher = { Roboflow },
                            year = { 2023 },
                            month = { apr },
                            note = { visited on 2025-12-30 },
                            }

@misc{
                            bicycle-zrn08_dataset,
                            title = { bicycle Dataset },
                            type = { Open Source Dataset },
                            author = { YOLOv8 },
                            howpublished = { \url{ https://universe.roboflow.com/yolov8-druip/bicycle-zrn08 } },
                            url = { https://universe.roboflow.com/yolov8-druip/bicycle-zrn08 },
                            journal = { Roboflow Universe },
                            publisher = { Roboflow },
                            year = { 2024 },
                            month = { apr },
                            note = { visited on 2025-12-30 },
                            }

@misc{
                            iranian-vehicle-cjfc5_dataset,
                            title = { iranian vehicle Dataset },
                            type = { Open Source Dataset },
                            author = { sipo },
                            howpublished = { \url{ https://universe.roboflow.com/sipo/iranian-vehicle-cjfc5 } },
                            url = { https://universe.roboflow.com/sipo/iranian-vehicle-cjfc5 },
                            journal = { Roboflow Universe },
                            publisher = { Roboflow },
                            year = { 2022 },
                            month = { aug },
                            note = { visited on 2025-12-30 },
                            }

@misc{
                            car-and-motorcycle-detection-with-kaggle-dataset_dataset,
                            title = { Car and Motorcycle Detection with Kaggle Dataset Dataset },
                            type = { Open Source Dataset },
                            author = { Aliff Haikal },
                            howpublished = { \url{ https://universe.roboflow.com/aliff-haikal-ssf6d/car-and-motorcycle-detection-with-kaggle-dataset } },
                            url = { https://universe.roboflow.com/aliff-haikal-ssf6d/car-and-motorcycle-detection-with-kaggle-dataset },
                            journal = { Roboflow Universe },
                            publisher = { Roboflow },
                            year = { 2022 },
                            month = { jun },
                            note = { visited on 2025-12-30 },
                            }

@misc{
                            signs-sqcaz_dataset,
                            title = { signs Dataset },
                            type = { Open Source Dataset },
                            author = { Bulat },
                            howpublished = { \url{ https://universe.roboflow.com/bulat-4geza/signs-sqcaz } },
                            url = { https://universe.roboflow.com/bulat-4geza/signs-sqcaz },
                            journal = { Roboflow Universe },
                            publisher = { Roboflow },
                            year = { 2023 },
                            month = { dec },
                            note = { visited on 2025-12-30 },
                            }

@misc{
                            bench-detector_dataset,
                            title = { Bench Detector Dataset },
                            type = { Open Source Dataset },
                            author = { labaml },
                            howpublished = { \url{ https://universe.roboflow.com/labaml/bench-detector } },
                            url = { https://universe.roboflow.com/labaml/bench-detector },
                            journal = { Roboflow Universe },
                            publisher = { Roboflow },
                            year = { 2023 },
                            month = { aug },
                            note = { visited on 2025-12-30 },
                            }

@misc{Reis2023,
  title         = {Real-Time Flying Object Detection with YOLOv8},
  author        = {Reis, Dillon and Kupec, Jordan and Hong, Jacqueline and Daoudi, Ahmad},
  year          = {2023},
  howpublished  = {arXiv preprint},
  eprint        = {2305.09972},
  archivePrefix = {arXiv},
  primaryClass  = {cs.CV},
  doi           = {10.48550/arXiv.2305.09972},
  url           = {https://arxiv.org/abs/2305.09972},
  note          = {Code: \url{https://github.com/dillonreis/Real-Time-Flying-Object-Detection_with_YOLOv8} (accessed Dec. 24, 2025)}
}

@article{Rahman2018,
  author  = {Rahman, Sami Ur and Ullah, Sana and Ullah, Sehat},
  title   = {Obstacle Detection in Indoor Environment for Visually Impaired Using Mobile Camera},
  journal = {Journal of Physics: Conference Series},
  year    = {2018},
  volume  = {960},
  number  = {1},
  pages   = {012046},
  doi     = {10.1088/1742-6596/960/1/012046}
}

@inproceedings{kasabekar2024smart,
  title={A Smart Device for Real-Time Assistance to the Visually Impaired People},
  author={Kasabekar, Om and Jeurkar, Shardul and Kamble, Riya and Srivel, R and Kamble, Milind},
  booktitle={2024 10th International Conference on Communication and Signal Processing (ICCSP)},
  pages={24--29},
  year={2024},
  organization={IEEE}
}
\vspace{5em}
\begin{IEEEbiography}[{\includegraphics[width=1in,height=1.25in,clip,keepaspectratio]{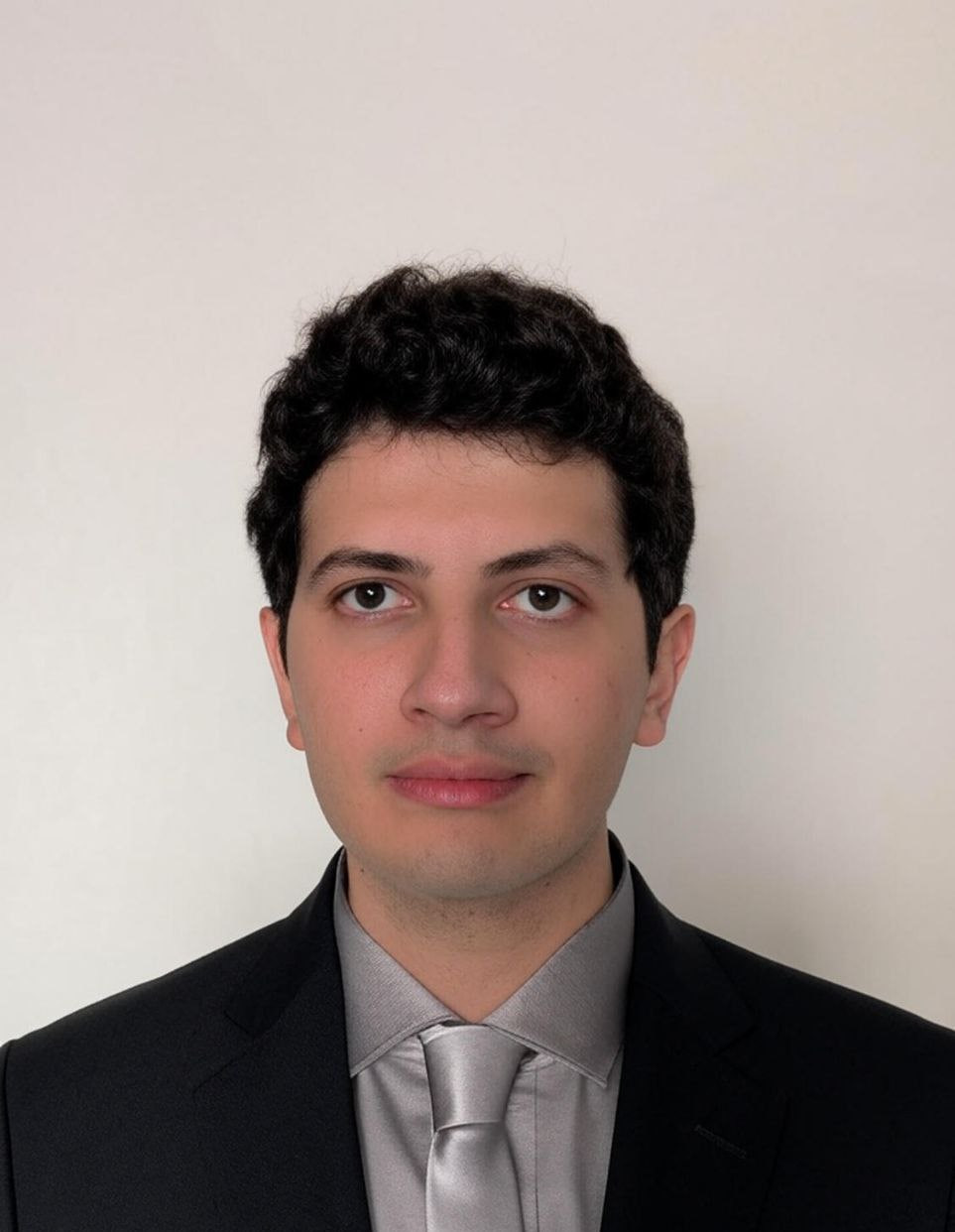}}]{Abtin Zandi}
received his B.Sc. degree in Computer Engineering from
Amirkabir University of Technology (Tehran Polytechnic), Tehran, Iran.
His research interests include computer vision, image generation, deep
learning, image processing, mobile computing, and real-time systems.
\end{IEEEbiography}

\begin{IEEEbiography}[{\includegraphics[width=1in,height=1.25in,clip,keepaspectratio]{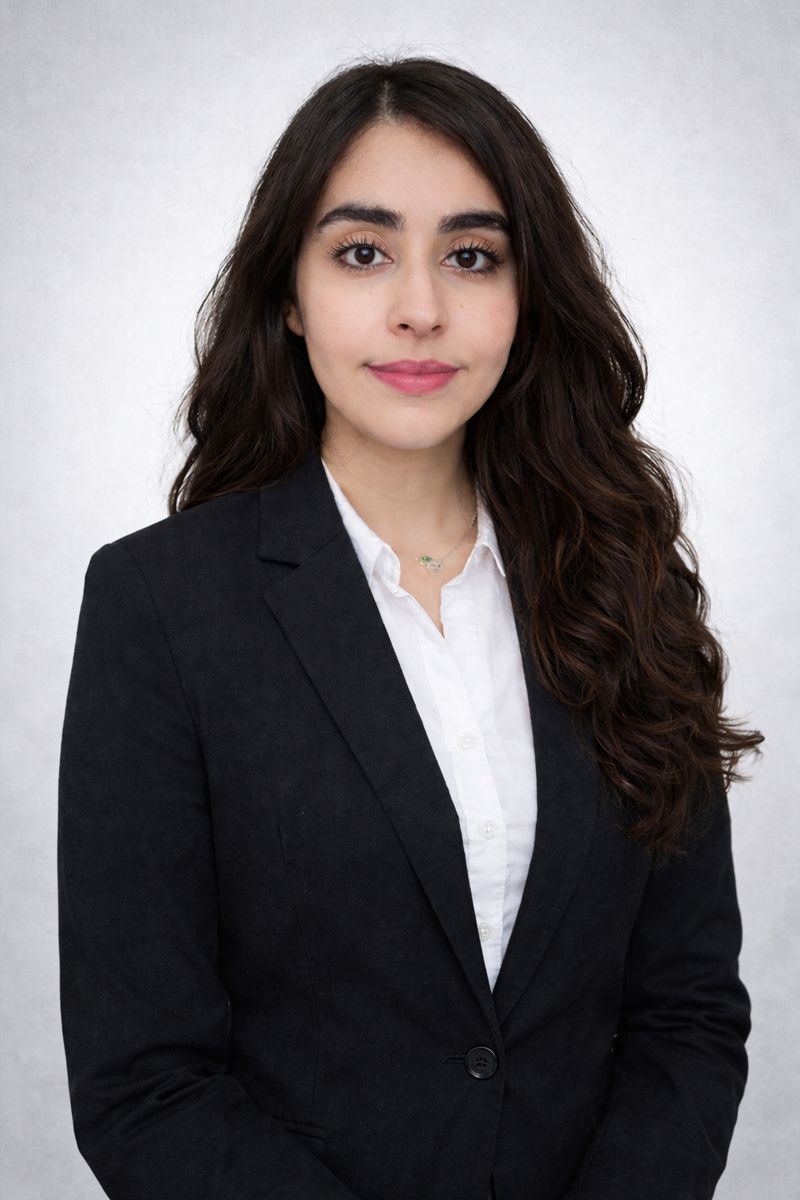}}]{Sarvin Nami}
received the B.Sc. degree in Computer Engineering at Amirkabir University of Technology (Tehran Polytechnic), Tehran, Iran. Her research interests include mobile programming, deep learning, multi-modal vision-language models, large language models (LLMs), and trustworthy AI.
\end{IEEEbiography}

\begin{IEEEbiography}[{\includegraphics[width=1in,height=1.25in,clip,keepaspectratio]{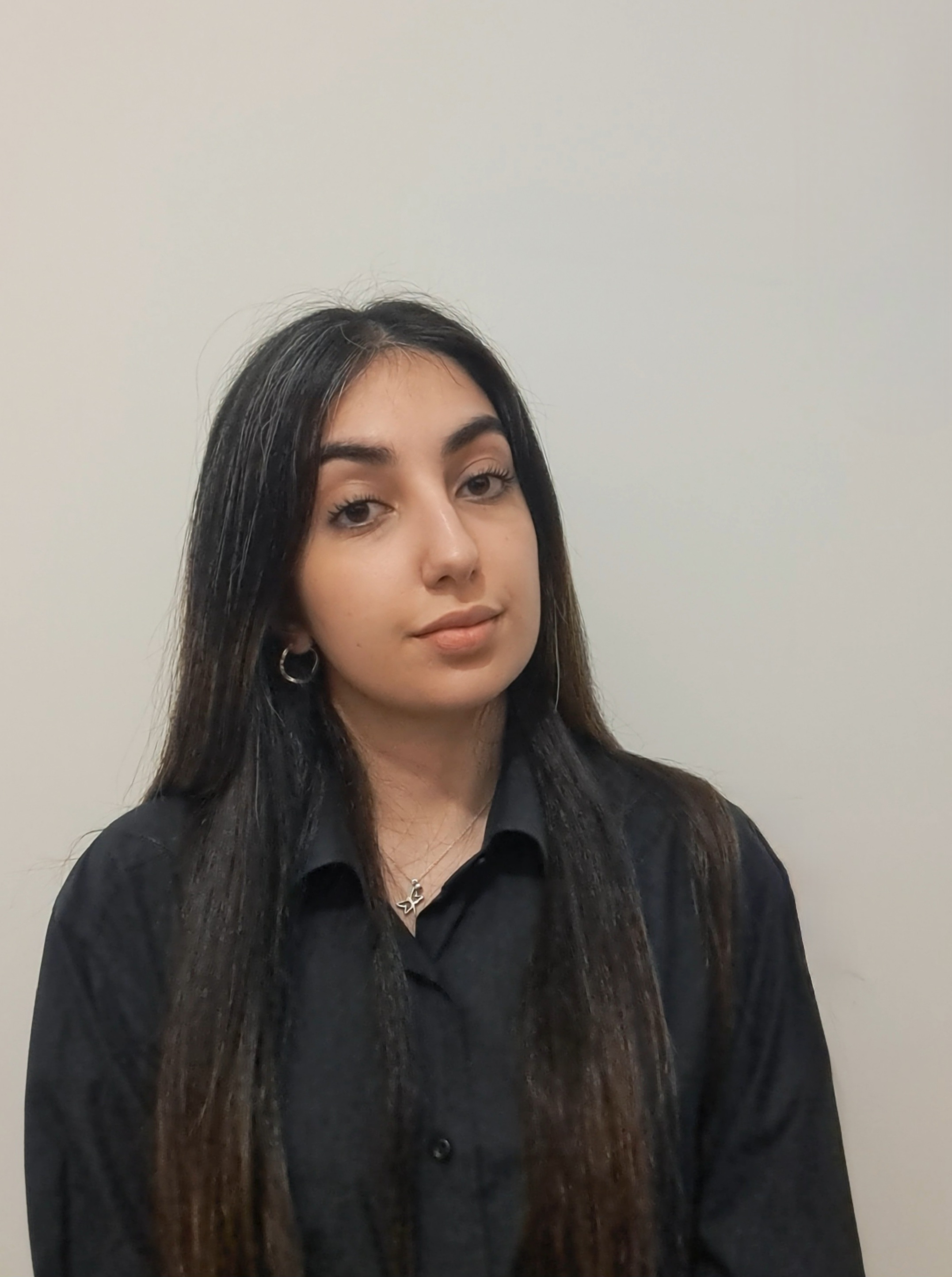}}]{Roza Ganjipour}
 is an undergraduate student in Computer Engineering at Amirkabir University of Technology (Tehran Polytechnic), Tehran, Iran. Her research interests include deep learning, computer vision, natural language processing (NLP), graph neural networks (GNNs), and large language models (LLMs).
\end{IEEEbiography}

\begin{IEEEbiography}[{\includegraphics[width=1in,height=1.25in,clip,keepaspectratio]{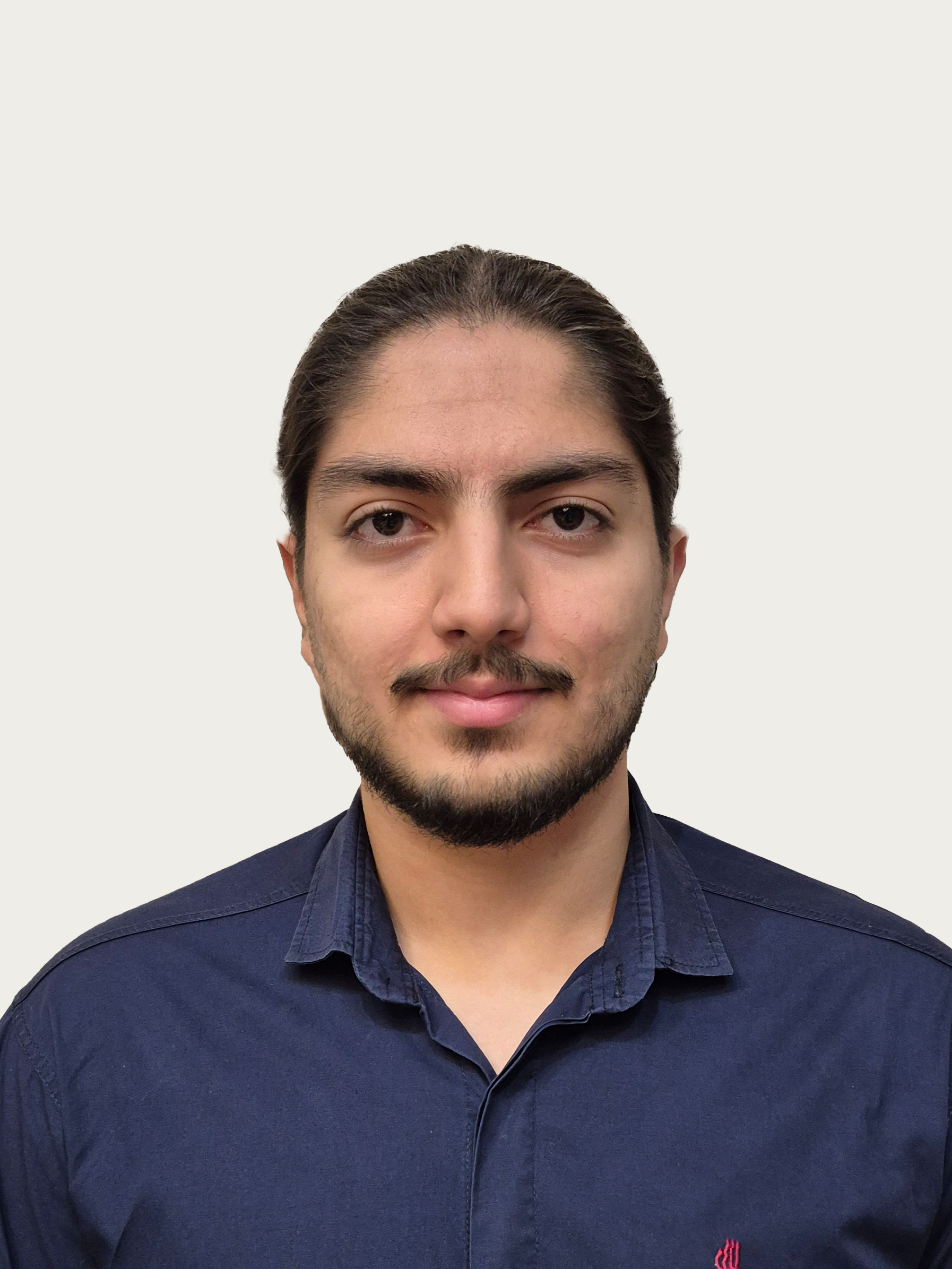}}]{Bardia Sabagh Kermani}
is an undergraduate student in Computer Engineering at Amirkabir University of Technology (Tehran Polytechnic), Tehran, Iran. His research interests include deep learning, computer vision, graph neural networks (GNNs), and real-time mobile and embedded intelligent systems.\end{IEEEbiography}

\begin{IEEEbiography}[{\includegraphics[width=1in,height=1.25in,clip,keepaspectratio]{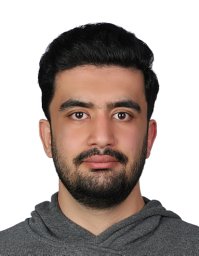}}]{Amirreza Tarabkhah}
received his B.Sc. degree in Computer Engineering from
Amirkabir University of Technology (Tehran Polytechnic), Tehran, Iran.
His research interests include computer vision, natural language processing, image processing.

\end{IEEEbiography}

\begin{IEEEbiography}[{\includegraphics[width=1in,height=1.25in,clip,keepaspectratio]{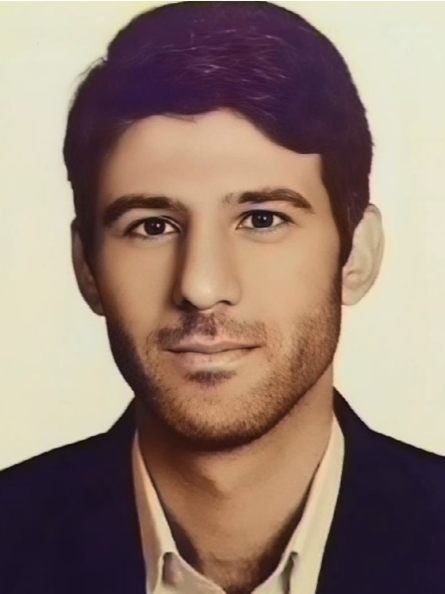}}]{Hamed Farbeh}
(Member, IEEE) received the B.Sc., M.Sc., and Ph.D. degrees in computer
engineering from the Sharif University of Technology (SUT), Tehran, Iran,
in 2009, 2011, and 2017, respectively. He was a member of the Dependable
Systems Laboratory (DSL), SUT, from 2007 to 2017; a Visiting Researcher
with the Embedded Computing Laboratory (ECL), KAIST, Daejeon, South Korea,
from October 2014 to May 2015; and a Postdoctoral Fellow with the Institute
of Research for Fundamental Sciences (IPM), Tehran, from May 2017 to
January 2018. He served as a faculty member in the Department
of Computer Engineering, Amirkabir University of Technology (Tehran
Polytechnic), Tehran, from 2017 to 2026, where he established the Intelligent Computing
and Communication Infrastructure Laboratory (ICCI) and served as Director
of the Computer Systems Architecture and Networks (CSAN) Group and Vice-Dean for Education. 
He is currently a Faculty Member of Al-Khwarizmi University, Urgench, Uzbekistan. His
current research interests include reliable memory hierarchy, emerging
memory technologies, AI processors, and cyber-physical systems.
\end{IEEEbiography}

\end{document}